%% file: main.tex
\RequirePackage{iftex}
\ifPDFTeX\pdfoutput=1\fi
\PassOptionsToPackage{dvipsnames}{xcolor}
\documentclass{agibot}

\input{packages}
\input{newcommands}

\title{Tracking Task Progress Through Errors and Recovery: A Benchmark and Recurrent Distributional Model}

\input{authors}
\input{sections/abstract}

\begin{document}
\maketitle

\input{sections/intro}
\input{sections/related_work}
\input{sections/method}
\input{sections/dataset}
\input{sections/exp}
\input{sections/conclusion}

\section*{Acknowledgment}
\AIDisclosure

\clearpage
\bibliographystyle{unsrt}
\bibliography{references}

\input{sections/appendix}
\end{document}

%% file: packages.tex
\usepackage{amsmath}
\usepackage{amssymb}

\usepackage{graphicx}
\graphicspath{{figures/}}

\usepackage{booktabs}
\usepackage{multirow}
\usepackage{etoolbox}
\usepackage{adjustbox}

\usepackage{siunitx}

\usepackage{xcolor}

\usepackage{url}

\usepackage[T1]{fontenc}
\usepackage{microtype}

%% file: newcommands.tex
\newcommand{\method}{REMORA}

\newcommand{\obs}{\mathbf{o}}

\newcommand{\mem}{\mathbf{M}}
\newcommand{\feat}{\mathbf{F}}

\newif\ifresultsready
\resultsreadyfalse

\newcommand{\AIDisclosure}{%
  OpenAI ChatGPT was used to polish the writing and assist plotting figures.
}

%% file: authors.tex
\renewcommand{\authorformat}[2][]{\authorfont\mbox{#2$^{#1}$}}
\author[1,*,\ddagger]{\textbf{Chen Wang}}
\author[4,\ddagger]{\textbf{Xinze Niu}}
\author[1,\ddagger]{\textbf{Zihao Zhu}}
\author[1,2]{\textbf{Keru Zhou}}
\author[3]{\textbf{Shihao Jin}}
\author[1,3]{\textbf{Yicheng Xu}}
\author[1]{\textbf{Wenhao Wang}}
\author[1]{\textbf{Bowen Jiang}}
\author[1,\dagger]{\textbf{Chang Zhou}}
\author[1]{\textbf{Li Liu}}
\author[1,*]{\textbf{Maoqing Yao}}

\affiliation[1]{Maniformer}
\affiliation[2]{Tsinghua University}
\affiliation[3]{Zhejiang University}
\affiliation[4]{Shanghai Jiao Tong University}

\newcommand{\PaperLogos}{%
  \includegraphics[height=12mm]{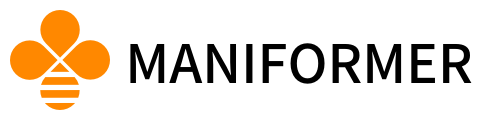}%
  \hfill\includegraphics[height=12mm]{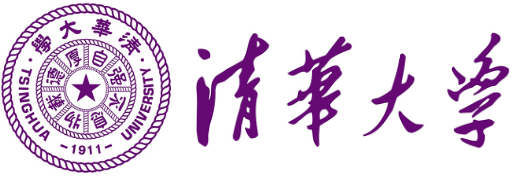}%
  \hspace{3mm}\includegraphics[height=12mm]{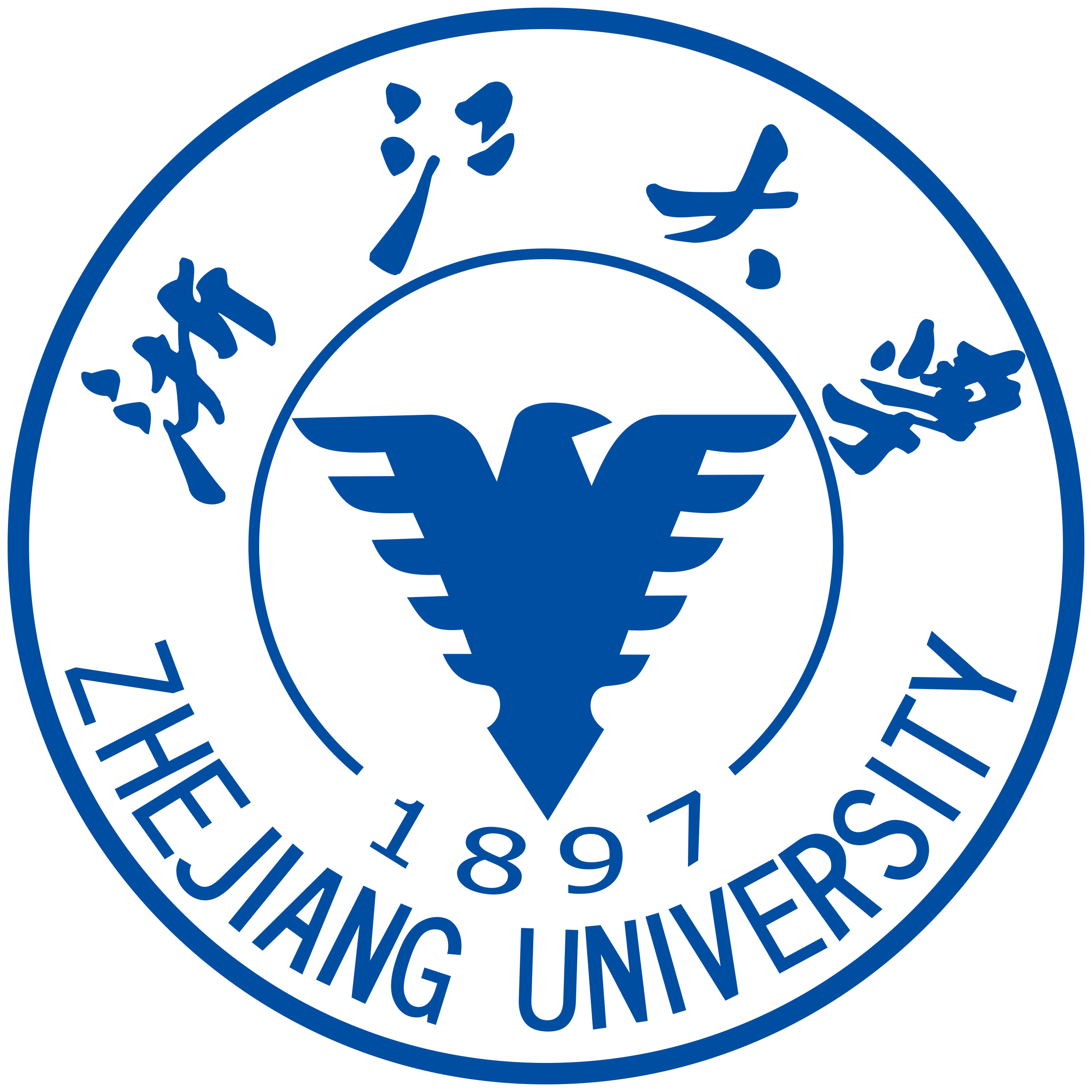}%
  \hspace{3mm}\includegraphics[height=12mm]{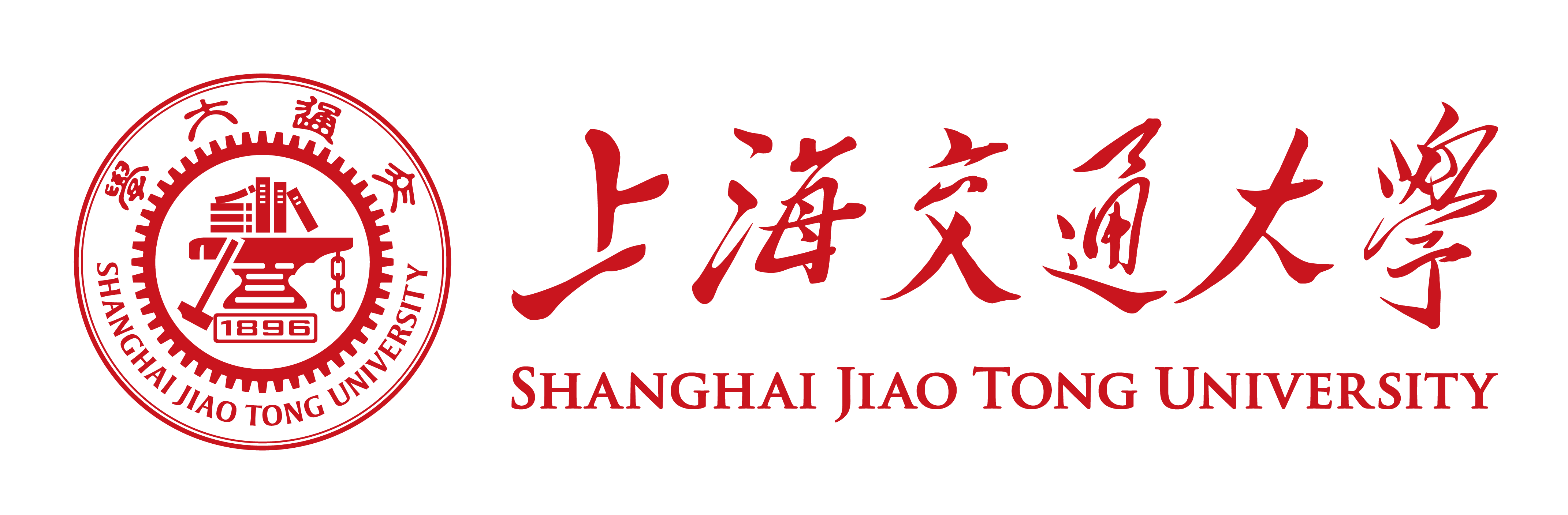}%
}
\fancypagestyle{firststyle}{%
  \fancyhf{}%
  \fancyhead[C]{\PaperLogos}%
  \fancyfoot[C]{\thepage}%
}

\makeatletter
\appto{\@thanks}{%
  \begingroup
    \footnotetext[1]{Corresponding authors: Chen Wang(wangchen2@maniformer.ai) and Maoqing Yao(yaomaoqing@agibot.com).}%
    \footnotetext[2]{Project leader: Chang Zhou.}%
    \footnotetext[3]{Equal contribution.}%
  \endgroup
}
\makeatother

%% file: sections/abstract.tex
\abstract{
Operational strategies for long-horizon robotic tasks require reliable evaluation
to determine whether execution is progressing, stalling, regressing, or recovering
from errors. Yet real-world robot datasets that capture error recovery and provide
frame-level error annotations remain scarce. To address this gap, we introduce
and open-source ReSTEP, a dataset of 3,156 episodes across two manipulation tasks
with semantic stage labels and frame-level error annotations. ReSTEP supports
training and benchmarking progress monitors through successful execution,
mistakes, and recovery. Existing progress estimators can struggle to distinguish
visually similar task stages and maintain temporally consistent predictions while
responding to genuine setbacks. We therefore propose \method{}, an online progress
estimator that combines recurrent visual memory, distributional prediction, and
causal peak selection. Memory retains execution history to help resolve visual
ambiguity, while distributional prediction represents alternative progress
hypotheses and causal peak selection tracks a supported hypothesis over time.
Together, these components enable consistent progress tracking that remains
responsive to errors and recovery. Experiments on both ReSTEP tasks demonstrate
that \method{} outperforms the state-of-the-art SARM baseline in progress accuracy
and temporal consistency.
}

%% file: sections/intro.tex
\section{Introduction}
\label{sec:introduction}

Recent advances in vision-language-action (VLA) models have enabled
language-conditioned robots to perform increasingly diverse manipulation tasks
and generalize to unfamiliar objects and
environments~\cite{brohan2023rt2,kim2024openvla,black2024pi0,black2025pi05}.
However, strong action-generation capabilities do not by themselves provide
reliable self-assessment: a policy may produce plausible motions while stalling
or undoing earlier work. This limitation becomes especially consequential in long-horizon tasks. A complementary task-progress estimator can provide
monitoring, changes in estimated task progress can be used to construct dense
rewards for reinforcement learning~\cite{zhang2026vlac,tan2026gprm} and to reweight
demonstration samples in imitation
learning~\cite{ma2025gvl,chen2026sarm,mao2026arm}. Progress feedback could also help address
the substantial data demands of VLA training by identifying stalled or regressing
executions that warrant expert corrections during DAgger-style data
collection~\cite{ross2011dagger,spencer2020interventions}. These applications
require progress estimates that remain temporally consistent under partial
observability while remaining sensitive to genuine setbacks.

Existing work on visual progress and reward estimation broadly follows two scoring designs. One derives scores from VLM embedding similarities~\cite{sontakke2023roboclip,rocamonde2024vlmrm} or judgments generated by pretrained or fine-tuned VLMs~\cite{ma2025gvl,zhang2026vlac,tan2026gprm,zhang2026progresslm,lee2026roboreward,wu2026lrm}. The other trains dedicated prediction heads or networks for progress, relative progress, or task completion, including heads built on pretrained VLM or VLA representations~\cite{becattini2020amidone,chen2026sarm,zhang2025rewind,mao2026arm,liang2026robometer}. These approaches demonstrate the utility of visual progress signals, with several reporting improvements in downstream robot learning. Nevertheless, limitations in spatial and temporal reasoning can make VLM-derived scores unreliable for subtle changes in task state~\cite{rocamonde2024vlmrm,zhang2026progresslm}. Coarse scoring scales can obscure small improvements~\cite{lee2026roboreward,wu2026lrm}, while time-derived supervision can conflate elapsed time with actual task advancement~\cite{zhang2025rewind,chen2026sarm,wu2026lrm,zhang2026progresslm}. Moreover, some methods rely on full-trajectory context or retrospective progress reconstruction, limiting their applicability to online progress monitoring~\cite{ma2025gvl,mao2026arm}.

\begin{figure}[t]
    \centering
    \includegraphics[width=0.75\linewidth]{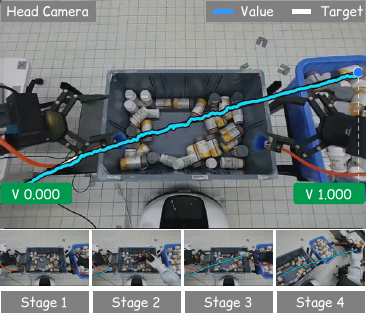}
    \caption{\textbf{Stage-aware progress supervision in a representative bottle-transfer episode.}
Head-camera observations are partitioned into four task stages, whose start frames are shown below the trajectory. The white curve denotes the normalized progress target and the cyan curve denotes the corresponding value estimate. Progress is anchored at $V=0$ at the episode start and $V=1$ upon task completion.}
    \label{Fig1}
\end{figure}

Beyond these challenges, progress estimation requires modeling temporal dependencies in both visual observations and progress predictions. On the input side, successive images reveal motion, state changes, and event order that may be ambiguous or unobservable in an isolated frame. Integrating this information helps distinguish visually similar observations associated with different task stages~\cite{becattini2020amidone}. However, short or sparsely sampled observation histories can omit earlier events needed to interpret the current state, while processing more frames increases computational cost~\cite{chen2026sarm,liang2026robometer}. On the output side, successive progress estimates should describe a coherent task evolution. Many methods ultimately produce a scalar estimate through direct regression or by taking the expectation of a predicted distribution~\cite{zhang2025rewind,chen2026sarm,liang2026robometer}. When the available visual history leaves multiple task stages plausible, the conditional progress distribution can remain multimodal. Squared-error regression or an expectation-based readout can then average distinct hypotheses into an intermediate value that corresponds to neither stage. 

Learning from unsuccessful executions presents a further challenge because dataset with error frame annotation are difficult to obtain. Methods such as ReWiND and SARM therefore augment demonstrations with temporally reversed segments to provide supervision for decreasing progress~\cite{zhang2025rewind,chen2026sarm}. Although useful, time reversal can produce physically implausible sequences and does not necessarily capture the dynamics of real mistakes and subsequent recovery~\cite{zhang2025rewind}. Restricting supervision to successful demonstrations and synthetic reversals can consequently leave informative failure and recovery behavior underused.

We introduce \method{}, a framework for online task-progress estimation that
combines temporal visual context with distributional prediction and causal peak
selection (Fig.~\ref{fig:overview}). We develop and study this framework on two
real-world logistics tasks using demonstrations that include successful
executions, mistakes, and recoveries. Comparison with SARM on parcel handling shows
that \method{} achieves lower mean absolute error and higher Spearman rank
correlation, indicating more accurate progress estimates and better preservation
of the relative progress ordering across observations.

Our contributions are threefold:
\begin{itemize}
  \item \textbf{ReSTEP: An error-annotated dataset for progress estimation.}
  We release more than 15 hours of robot demonstrations across 3,156 episodes
  covering parcel handling and medicine-bottle scanning and storage. Semantic
  stage annotations and frame-level error labels capture successful executions,
  mistakes, and recoveries, supporting both model training and benchmarking.

  \item \textbf{A recurrent distributional progress model.}
  We augment a pretrained VLA backbone with categorical value heads and a
  recurrent visual memory to represent ambiguous progress states in their
  temporal context. Stage and error annotations supervise both advancing and
  decreasing progress, retaining observations of mistakes and recoveries
  during training.

  \item \textbf{Causal peak selection for temporal consistency.}
  We introduce a decoder that selects a supported peak from the current progress
  distribution based on the previous estimate. This rule favors continuity
  across successive predictions while allowing decreases supported by the
  distribution, providing a causal readout of potentially multimodal progress.
\end{itemize}
%

%% file: sections/related_work.tex
\section{Related Work}
\label{sec:related}

\subsection{Inverse reinforcement learning}

Inverse reinforcement learning (IRL) is closely related to progress and visual
reward modeling: both learn evaluative signals from data to reduce manual reward
design. IRL seeks rewards that explain demonstrated behavior and can guide policy
learning toward similar behavior~\cite{ng2000airl}. Early feature-based IRL
methods share the principle of matching demonstration statistics: apprenticeship
learning seeks policies with matching feature
expectations~\cite{abbeel2004alirl}, while maximum-entropy IRL selects the
highest-entropy trajectory distribution consistent with these
constraints~\cite{ziebart2008meirl}. Later neural approaches extend this
foundation: Guided Cost Learning combines maximum-entropy learning of neural
costs with sample-based policy optimization~\cite{finn2016gcl}, while GAIL
connects regularized IRL to adversarial imitation by matching expert and learner
state--action visitation distributions~\cite{ho2016gail}. These lines of work
overlap, but differ in their primary objective: IRL explains demonstrated
behavior through rewards, whereas progress and visual reward models directly
score task completion or advancement from observations. Such scores can assess
execution without requiring the evaluated policy to reproduce the
demonstrator's behavior.

\subsection{Progress and visual reward models}

Progress and visual reward models convert observations into scores of task
advancement, providing feedback for policy learning and execution monitoring.
Early progress prediction uses recurrent video
models~\cite{becattini2020amidone}; temporal cycle consistency learns frame
alignments~\cite{dwibedi2019tcc}, which XIRL uses to construct goal-reaching
rewards across embodiments~\cite{zakka2022xirl}. Subsequent work pursues both
transfer across tasks and precise estimation within specified tasks.

Multi-task approaches seek reusable reward functions across language-specified
goals. LIV and RoboCLIP exploit vision--language representations for reward
prediction~\cite{ma2023liv,sontakke2023roboclip}, while VLC and ReWiND incorporate
temporal ranking and progress supervision, respectively, to improve
transfer~\cite{alakuijala2025vlc,zhang2025rewind}. Building on broader VLM
capabilities, GVL and ProgressLM infer progress using visual context or reference
demonstrations~\cite{ma2025gvl,zhang2026progresslm}. RoboReward and Large Reward
Models scale supervised outcome and progress scoring across diverse
tasks~\cite{lee2026roboreward,wu2026lrm}, while Robometer adds trajectory
comparisons to learn from unsuccessful executions~\cite{liang2026robometer}.

Task-level approaches fit evaluation to a specific task. RL-VLM-F learns a
separate reward model from VLM preferences for each task~\cite{wang2024rlvlmf},
while SARM uses procedure-specific stage annotations to account for variations
in execution timing~\cite{chen2026sarm}. These directions are complementary:
Robo-Dopamine and ProcVLM combine broad pretraining with task-specific
adaptation~\cite{tan2026gprm,feng2026procvlm}.

Our work adopts the task-level setting because reliable temporal estimation
remains challenging even for familiar procedures: visually similar states can
correspond to different stages, and mistakes can interrupt or reverse
progress~\cite{chen2026sarm,tan2026gprm}. Broad task coverage alone does not
establish accurate, temporally coherent monitoring. Distributional models
already represent progress, remaining steps, or returns, but use expected-value
readouts~\cite{liang2026robometer,seyedghasemipour2025siefm,physicalintelligence2025pi06}.
Task-specific stage and mistake annotations provide a controlled basis for
studying these unresolved estimation problems. Within this setting, we combine
recurrent visual memory with causal peak selection to track a supported
progress hypothesis through mistakes and recovery.

%% file: sections/method.tex
\section{Method}
\label{sec:method}

\input{figures/overview}

\subsection{Problem formulation and overview}

At time $t$, the estimator receives multi-camera observations $\obs_t$,
robot state $s_t$, and a language instruction $\ell$.
It estimates normalized task progress using observations available
through $t$.
During training, semantic stage boundaries and annotated error intervals
define a reference target $\tilde y_t\in[0,1]$.
We model the corresponding progress distribution as
\begin{equation}
  p_t(k)
  =
  p_\theta\!\left(
    \tilde Y_t=z_k \mid \obs_{\leq t},s_t,\ell
  \right),
  \label{eq:value}
\end{equation}
where $\{z_k\}_{k=0}^{K-1}$ are fixed progress atoms spanning $[0,1]$.
Throughout the paper, ``value'' refers to annotation-defined task progress.

Fig~\ref{fig:overview} summarizes the training and online
inference pipeline.
Multi-camera images are encoded by pretrained vision encoder SigLIP~\cite{zhai2023sigmoid} and enriched with execution
history through recurrent memory fusion.
The VLA backbone combines these visual features with the language
instruction and robot state, and a categorical value head reads out
the action-expert suffix conditioned on all-zero action input to predict the progress distribution $p_t$.
During training, semantic stage boundaries and annotated error
intervals define progress targets, which are encoded as soft labels
to supervise this distribution.
During online inference, progress tracking selects a supported peak
according to the previous estimate $\hat y_{t-1}$, yielding the
current estimate $\hat y_t$.
Visual memory thus supplies historical context for distribution
prediction, while peak association maintains continuity in the
scalar readout.
Neither stage nor error annotations are required at inference.

\subsection{Error-aware progress targets}
\label{sec:mistake-target}

We construct progress targets from semantic stage annotations and localized execution errors. Each stage is assigned a fixed progress mass,
shared across demonstrations of the same task, with the task-specific
weights specified in the implementation details. Stages omitted because
their requirements are already satisfied contribute initial progress
credit. For stage $j$ spanning frames $[b_j,d_j)$, let $B_j$ and
$B_{j+1}$ denote its prescribed progress boundaries. In the absence of
errors, the target increases linearly between these boundaries.
This construction aligns demonstrations at semantic stage endpoints
while allowing their execution durations to vary.

To retain supervision from real mistakes, we replace continued growth
within annotated error intervals with a bounded decrease. For an error
interval of length $L$ ending at frame $c$, let $v$ be progress immediately before entering the error interval and $\alpha$ the preceding positive slope.
The target decreases by $\beta$ per frame during the error and then
increases with a recomputed slope $\alpha^{+}$:
\begin{equation}
\begin{aligned}
  \beta
  &= \min\!\left(\alpha,\frac{v}{L}\right),\\
  \alpha^{+}
  &= \frac{B_{j+1}-(v-\beta L)}{d_j-c},
  \qquad c<d_j.
\end{aligned}
\label{eq:error-descent}
\end{equation}
The cap keeps progress nonnegative and limits the descent rate to the
preceding rate of advancement. After each error, the positive slope is
recomputed toward the stage endpoint and applied until the next error
or stage boundary. These targets provide an annotation-based proxy for
setbacks and recovery, preserving mistake observations for value learning.
The annotations are used only to construct offline supervision and are
not required during online inference.

\subsection{Categorical prediction and online peak association}
\label{sec:categorical}

\paragraph{Categorical progress prediction}
Visually similar observations can correspond to different task stages.
We therefore predict a categorical progress distribution that can
retain multiple plausible hypotheses.
Target annotated in \ref{sec:mistake-target} are encoded as soft labels over neighboring progress
bins, which means each progress target from \ref{sec:mistake-target} is assigned to its two neighboring bins using linear interpolation to reduce inaccuracy introduced by discretizations. The prediction heads are trained with cross-entropy.

\paragraph{Online peak association}
Independently selecting the dominant peak can cause predictions to
jump between competing hypotheses.
We instead select the supported peak closest to the previous estimate:
\begin{equation}
  \hat k_t
  =
  \arg\min_{k\in\mathcal P_t}
  |z_k-\hat y_{t-1}|,
  \qquad
  \hat y_t=z_{\hat k_t},
  \label{eq:decoder}
\end{equation}
where $\mathcal P_t$ indexes the candidate peaks extracted from $p_t$.
This association favors temporal continuity while allowing progress
to increase or decrease.

\subsection{Gated recurrent visual memory}
\label{sec:memory}

\begin{figure}[!htbp]
  \centering
  \adjustbox{trim=0 30 0 0, clip, width=0.55\linewidth}{%
    \includegraphics{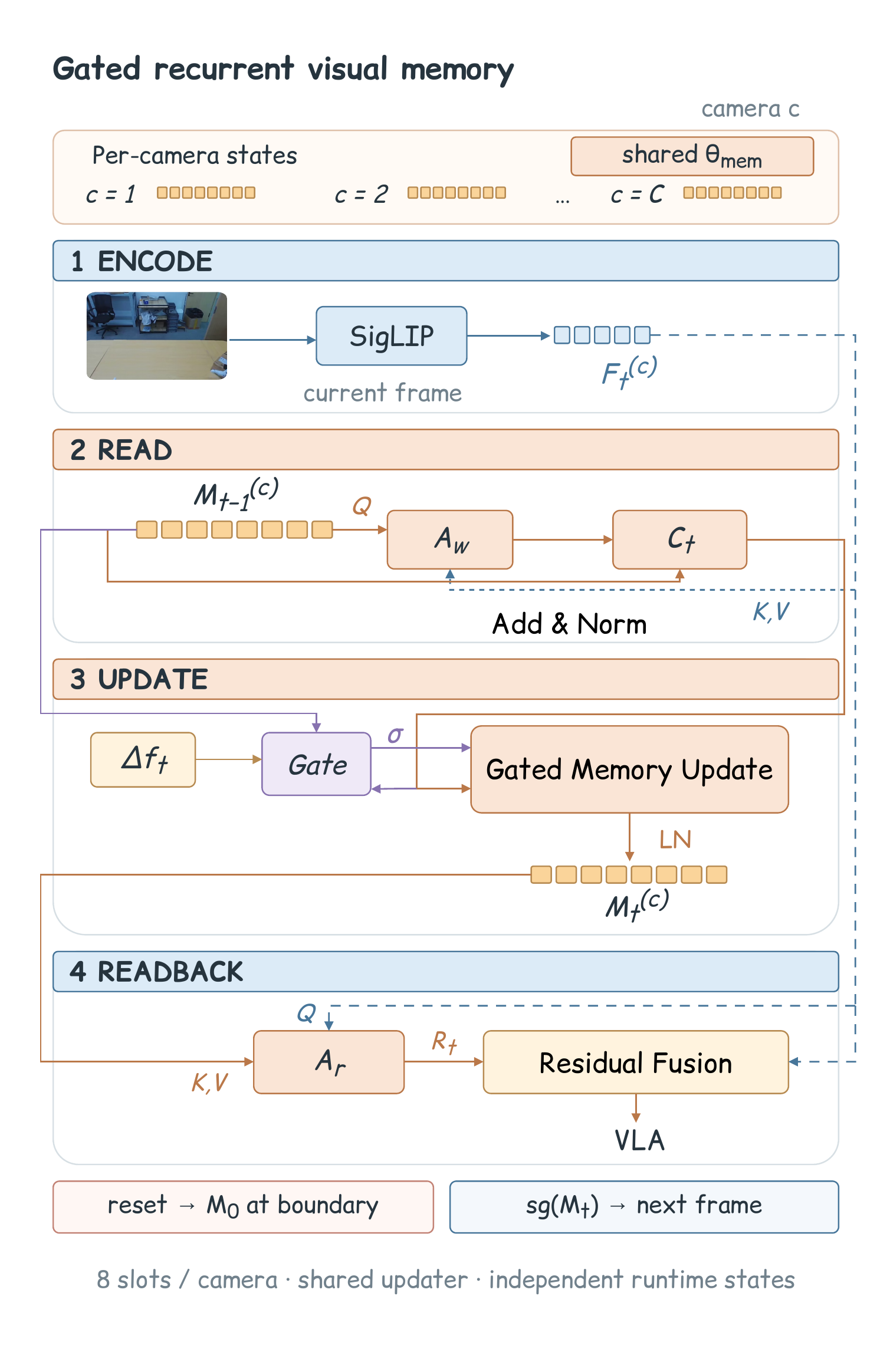}%
  }
    \caption{\textbf{Gated recurrent visual memory.}
    Each camera maintains eight memory slots, with independent states
    and shared update parameters.
    Memory queries current visual tokens to form a candidate state,
    and a frame-gap-conditioned gate updates the memory.
    Current visual tokens then attend to the updated memory for
    residual fusion.
    The updated state is carried to the next frame with stopped
    gradients during training and reset at stream boundaries.}
  \label{fig:memory-architecture}
\end{figure}

To incorporate execution history into progress estimation,
we recurrently update a fixed-size visual memory as new frames arrive.
At each step, the previous memory and current visual features jointly
determine the updated state, which enriches the current frame
representation and is carried forward to the next step.
This provides temporal context before the progress distribution is
predicted, complementing the online peak association that subsequently
selects among its modes.
As shown in Fig~\ref{fig:memory-architecture}, each of the $C$ cameras
maintains eight memory slots, with independent runtime states and
shared parameters $\theta_{\mathrm{mem}}$.
For camera $c$, the visual encoder
produces projected patch tokens $\feat_t^{(c)}$ from the current frame.
We omit the camera superscript below for clarity.

In the \textit{Read} step, cross-attention $A_w$ uses the previous
memory $\mem_{t-1}$ as queries and the current patch tokens as keys
and values, denoted by $Q$, $K$, and $V$ in Fig.~\ref{fig:memory-architecture}.
Its output is added to the previous memory and normalized to form
the candidate state $\mathbf C_t$, corresponding to
\textit{Add \& Norm}.

In the \textit{Update} step, the Gate network combines the previous
memory, the candidate state, and an encoding of the frame gap
$\Delta f_t$.
The sigmoid function $\sigma$ converts its output into elementwise
update weights $\mathbf G_t\in[0,1]$:
\begin{equation}
  \mem_t
  =
  \operatorname{LN}\!\left(
    (1-\mathbf G_t)\odot\mem_{t-1}
    +\mathbf G_t\odot\mathbf C_t
  \right).
  \label{eq:memory}
\end{equation}
Here, $\operatorname{LN}$ denotes layer normalization and $\odot$
denotes elementwise multiplication.
Small gate values favor retaining previous features, while large
values favor incorporating the candidate state.

In the \textit{Readback} step, cross-attention $A_r$ reverses the
attention direction: current patch tokens provide the queries,
and updated memory provides the keys and values.
The resulting readout $\mathbf R_t$ passes through a learned output
projection and is added to the current patch tokens by
\textit{Residual Fusion}, supplying the VLA predictor with temporal
context.
Between frames, $\operatorname{sg}(\mem_t)$ carries the updated state
forward, where $\operatorname{sg}$ denotes stop-gradient during
training.
At reset boundaries, all camera memories return to the learned
initial state $\mem_0$, and the decoder's previous progress estimate
is cleared jointly.

%% file: figures/overview.tex
\begin{figure}[t]
  \centering
  \includegraphics[width=\linewidth]{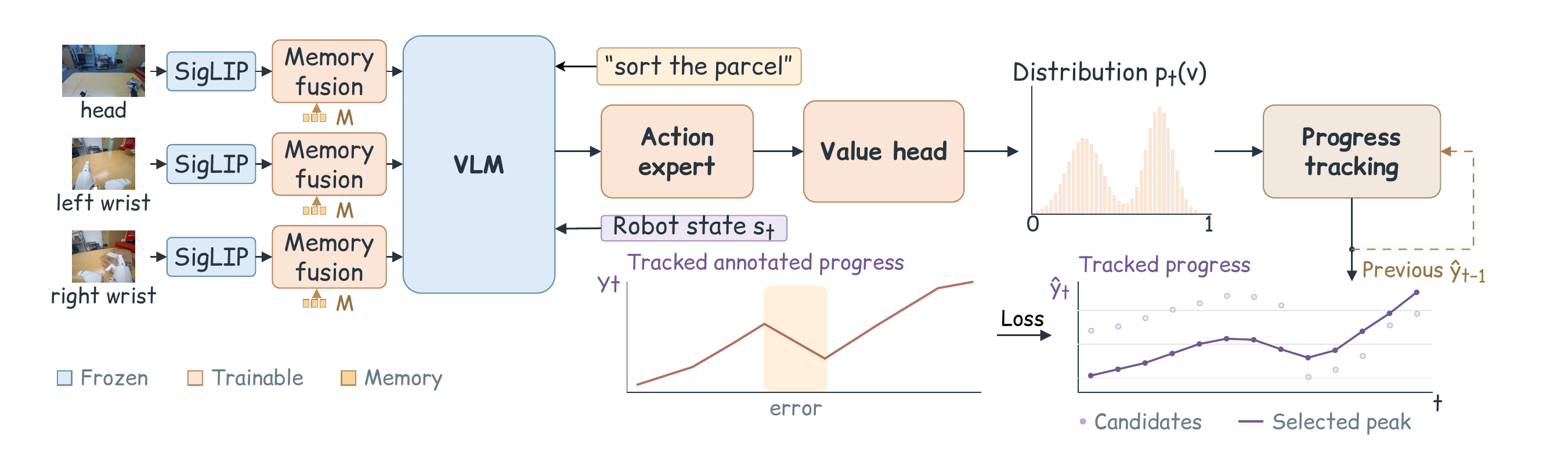}
  \caption{\textbf{Overview of \method{}.}
  Recurrent memory enriches multi-camera visual features before the
  VLA backbone and categorical value head predict a progress distribution.
  During training, stage and error annotations provide soft progress
  targets for distributional supervision.
  During online inference, progress tracking selects a supported peak
  using the previous estimate.
  Colors indicate frozen and trainable modules in the
  frozen-backbone configuration.}
  \label{fig:overview}
\end{figure}

%% file: sections/dataset.tex
\section{ReSTEP: A Dataset for Error-Aware Task Progress Estimation}
\label{sec:dataset}

We introduce ReSTEP (Real-Robot Stage-Annotated Trajectories for Error-Aware
Progress), a dataset covering two practical manipulation scenarios: parcel
handling (\textit{D1}) and medicine-bottle scanning and storage (\textit{D2}).
It contains 3,156 episodes and 1,692,390 synchronized frames, corresponding to
approximately \SI{15.7}{h} of robot operation.
The dataset includes expert demonstrations and autonomous executions with
human-in-the-loop corrections. We distinguish \emph{mistake-free episodes},
which contain no execution errors, from \emph{mistake episodes}, which contain
errors followed by recovery. Both types end in successful task completion.
Semantic stage annotations identify task phases, while
frame-level error annotations localize execution mistakes.

We will publicly release ReSTEP, containing the trajectories and their stage and error annotations to support progress-model training and evaluation. Our dataset also contains full observation and action information, which can also be used to train vision-language-action (VLA) or world-action (WA) models. 

\subsection{Tasks and Data Collection}

All trajectories were collected on an AgiBot G02 robot and contain synchronized
head and bilateral wrist RGB observations, proprioception, and actions at
\SI{30}{Hz}. Expert demonstrations were acquired through teleoperation. These
demonstrations also include episodes with execution mistakes, as errors can
occur during teleoperation. The bottle data additionally include
human-in-the-loop (HIL) collection, in which a policy acts autonomously and an
operator intervenes when correction is needed.
With the inclusion of HIL trajectories, \textit{D2} contains a higher proportion
of episodes with execution mistakes than \textit{D1}. Table~\ref{tab:dataset-resource} summarizes the statistics for \textit{D1} and
\textit{D2}.

\paragraph{\textit{D1}: Parcel handling}
The parcel data contain 1,953 expert-demonstration episodes across eight
scenarios, combining package type (box or bag), label orientation (up or down),
and initial placement (centered or off-center). Using dexterous hands, the robot
centers the parcel when necessary, flips it to expose the label, pushes it away,
and returns to its initial pose. The corresponding semantic stages are
\emph{move}, \emph{flip}, \emph{push}, and \emph{return}. Centered parcels omit
the move stage, and parcels with an upward-facing label omit the flip stage,
yielding four stage sequences with two to four stages per episode. These
variations allow progress estimation to be studied under different initial
conditions and task-stage sequences.

\paragraph{\textit{D2}: Medicine-bottle scanning and storage}
The bottle data contain 959 expert-demonstration episodes and 244 HIL episodes.
Using suction-equipped grippers, the robot grasps a bottle from a front bin,
rotates it to expose its QR code to the head-mounted camera, places it in a
neighboring basket, and resets its arms and body. Each episode has four ordered
stages: \emph{grasp}, \emph{scan}, \emph{place}, and \emph{reset}. HIL
trajectories include autonomous behavior and corrective actions within the same
execution, providing observations of mistakes and subsequent recovery.

\subsection{Stage and Error Annotations}

Each stage annotation consists of a semantic label and a half-open frame interval
$[s_k,e_k)$. Stage annotations cover 584,348 frames in parcel handling and
699,433 frames in bottle scanning. They identify the task phase associated with
an observation and provide anchors for constructing progress supervision across
episodes of different durations. Stage annotations span the interval from task
initiation to completion. The recordings also include frames before task
initiation and after completion, so the number of stage-annotated frames is
smaller than the total number of recorded frames. Separate \emph{Error Frame} annotations localize execution errors, including empty grasps and object drops. Parcel handling contains 18 error intervals
covering 461 frames, and bottle scanning contains 400 intervals covering
22,002 frames. 

The stage labels and error intervals provide annotation-level information for
reuse with different progress estimators. Our method converts these annotations
into the advance--decline--recovery targets defined in
Section~\ref{sec:mistake-target}. The resulting continuous progress values are
constructed supervision; the underlying annotations remain available for
alternative target definitions and analyses of within-stage progress, local
errors, and corrective behavior.

\begin{table}[t]
  \centering
  \caption{ReSTEP composition by task and collection source. Mistake episodes
  contain an available mistake annotation; all episodes end in successful task
  completion.}
  \label{tab:dataset-resource}
  \small
  \setlength{\tabcolsep}{4pt}
  \begin{tabular}{@{}lrrr@{}}
    \toprule
    Source & Episodes & Frames & \shortstack{Mistake\\episodes} \\
    \midrule
    \textit{D1} (expert) & 1,953 & 793,632 & 18 \\
    \textit{D2} (expert) & 959 & 657,159 & 66 \\
    \textit{D2} (HIL) & 244 & 241,599 & 179 \\
    \midrule
    Total & 3,156 & 1,692,390 & 263 \\
    \bottomrule
  \end{tabular}
\end{table}

%% file: sections/exp.tex
\section{Experiments}
\label{sec:experiments}

We evaluate \method{} on the ReSTEP dataset
(Section~\ref{sec:dataset}) to address four questions:
\begin{enumerate}
  \item[\textbf{Q1:}] Does our recurrent memory design improve progress estimation?
  \item[\textbf{Q2:}] How does freezing or fine-tuning the VLM
  backbone affect performance?
  \item[\textbf{Q3:}] How does causal peak selection affect prediction accuracy
  and temporal consistency?
  \item[\textbf{Q4:}] How does our method compare with SARM, a state-of-the-art
  progress estimation method?
\end{enumerate}

\subsection{Implementation Details}

\textbf{Data preparation.}
All splits are disjoint at the episode level and shared across models.  Parcel handling (\textit{D1}) uses 1,757 training and 196 test episodes.  For bottle scanning (\textit{D2}), we split the expert and human-in-the-loop sources separately and hold out approximately 10\% of each group for testing. This yields 1,081 training, and 120 test episodes. The test sets contain one mistake episode in \textit{D1} and 25 in \textit{D2}.Stage-duration weights are estimated from mistake-free training episodes. Stages already satisfied at the start of an instruction contribute initial progress. Error frames remain in training and receive the descent targets defined in Section~\ref{sec:mistake-target}.

\textbf{Model and training.}
We implement \method{} on a $\pi_0$-style backbone with a categorical value head predicting 201 uniformly spaced progress bins in $[0,1]$.  The model receives multi-view RGB observations, proprioception, and a language
instruction through the state-only readout described in Section~\ref{sec:method}. All variants within a task start from the same behavior-cloning checkpoint and
are fine-tuned using the value-only objective with 30,000 steps. Training is performed on eight NVIDIA H200 GPUs. Recurrent training processes observations in temporal order and uses gradient accumulation to match the effective batch size of the window-based models. All variants share the same progress targets and normalization statistics. \footnote{Code will be publicly released upon acceptance.}

\begin{figure}[t]
  \centering
  \includegraphics[width=0.75\linewidth]{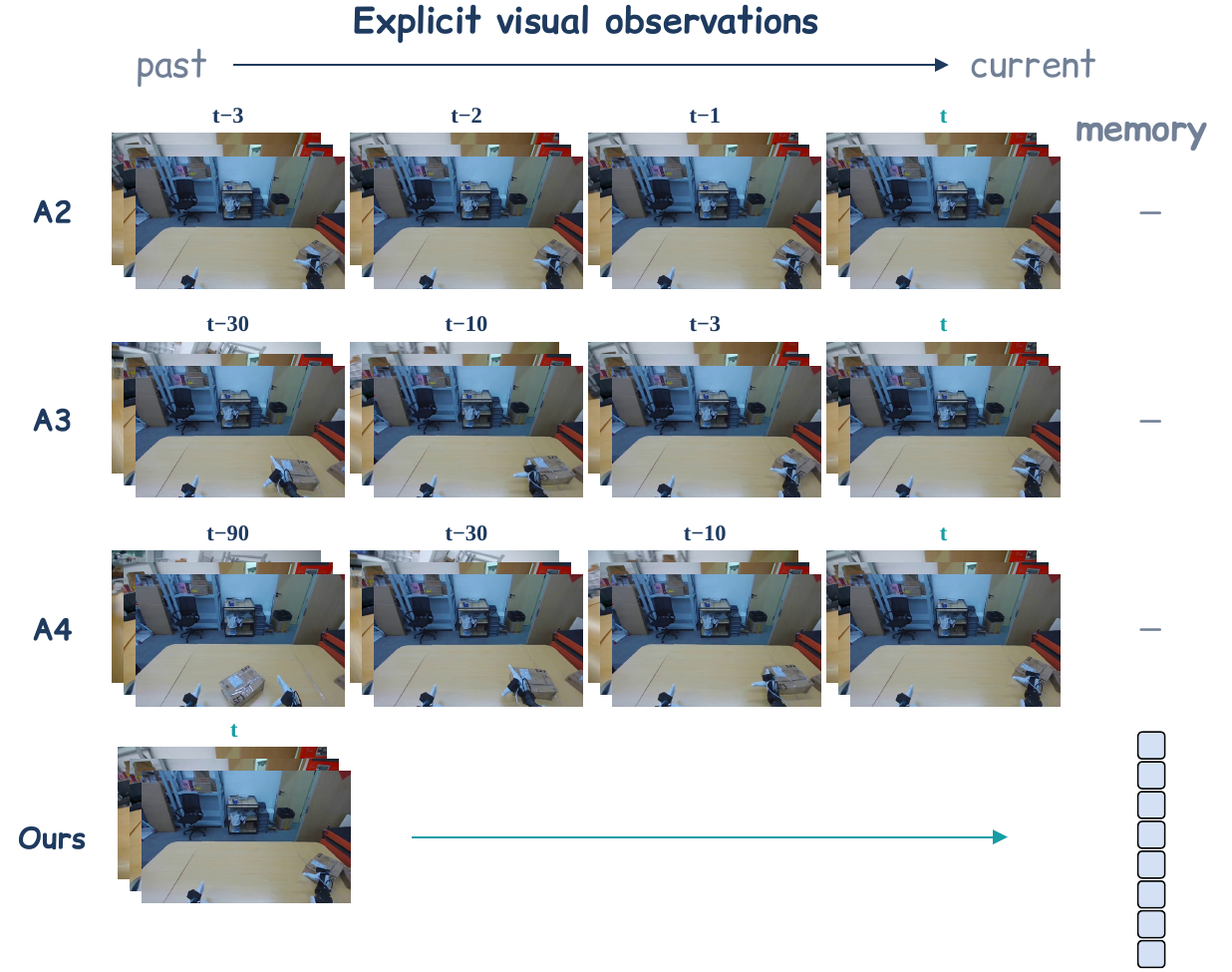}
  \caption{\textbf{Temporal input configurations.}
  A2 - A4  use four explicit multi-view observations sampled at $(t-3,t-2,t-1,t)$, $(t-30,t-10,t-3,t)$ and $(t-90,t-30,t-10,t)$,
  respectively. \method{} uses only the current observation and retains temporal
  context through eight gated recurrent memory tokens.}
  \label{fig:temporal-context}
\end{figure}

\subsection{Metrics}
We evaluate progress accuracy, ordering, and temporal smoothness using four
metrics. All metrics use the evaluated predictions without additional smoothing, interpolation, or clipping. Let $\mathcal E$ denote
the set of test episodes and $\mathcal F_e$ the evaluated frame indices within
the annotated instruction spans of episode $e$, including mistake frames.
At frame $t$, $\hat y_{e,t}$ and $\tilde y_{e,t}$ denote the predicted and
reference progress, respectively.

\textbf{Progress accuracy.}
Mean absolute error (MAE) $A$ measures the discrepancy between predicted and reference progress:
\begin{equation}
  A
  = \frac{1}{|\mathcal E|}\sum_{e\in\mathcal E}
    \frac{1}{|\mathcal F_e|}\sum_{t\in\mathcal F_e}
    \left|\hat y_{e,t}-\tilde y_{e,t}\right|.
  \label{eq:point-metrics}
\end{equation}
Lower $A$ indicates more accurate estimates on the normalized $[0,1]$ scale.

\begin{figure}[t]
  \centering
  \adjustbox{trim=0 10 0 0, clip, width=\linewidth}{%
    \includegraphics{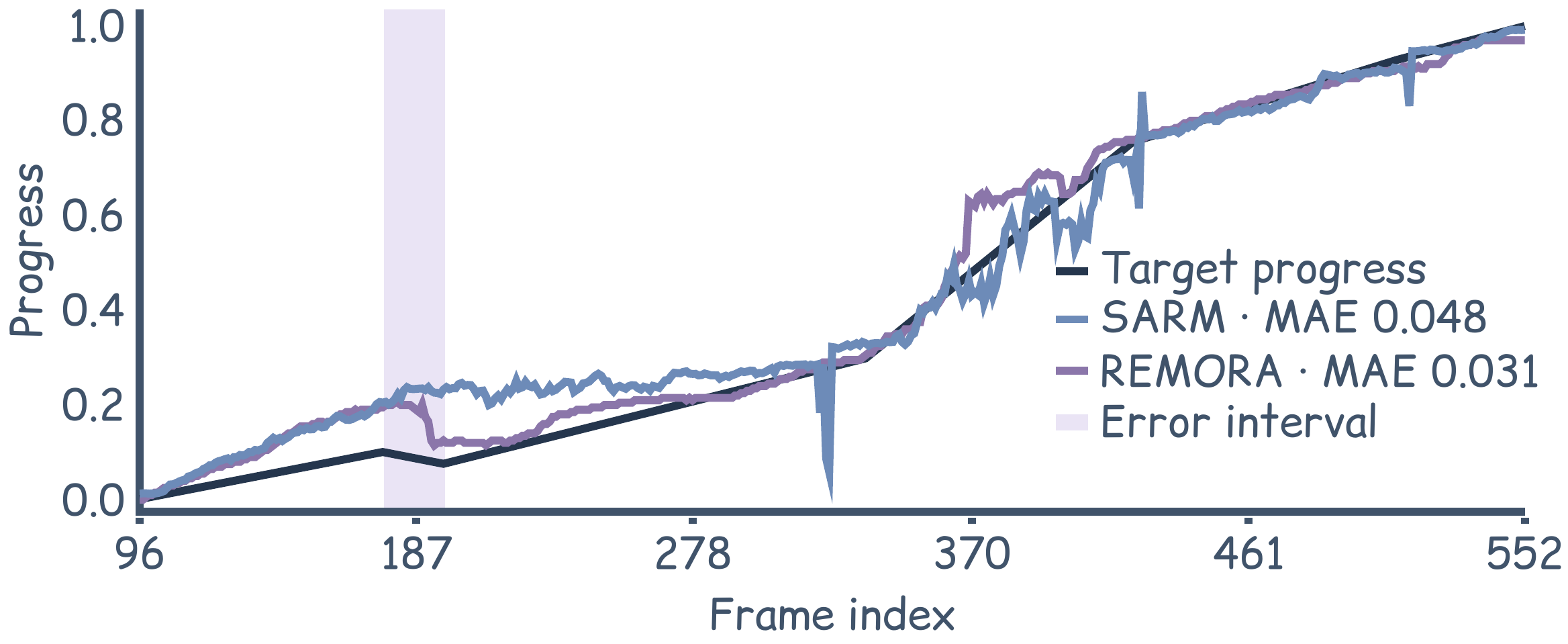}%
  }
  \caption{\textbf{Representative progress estimation on a D2 trajectory.}
  The black curve denotes the stage- and error-aware target, while the shaded region indicates an annotated error interval. \method{} better captures the target decrease and subsequent recovery than SARM, achieving a lower episode-level MAE.}
  \label{fig:d2-baseline-trajectory}
\end{figure}

\begin{figure}[!htbp]
  \centering
  \includegraphics[width=\linewidth]{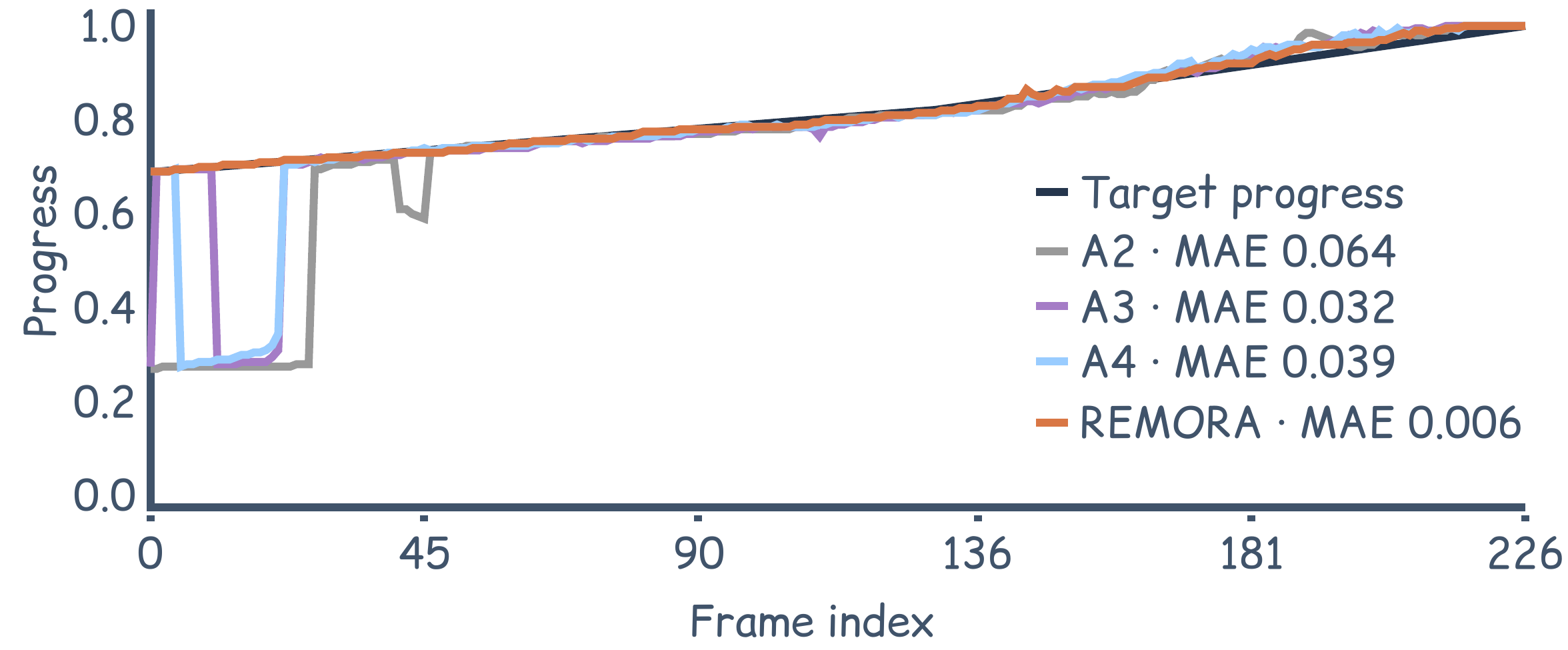}
  \caption{\textbf{Representative progress estimation on a D1 trajectory.}
  The black curve denotes the target progress. The
  \method{} accurately tracks the nonzero initial target and subsequent progression, whereas A2 - A4 exhibit transient underestimation at the instruction onset. \method{} achieves the lowest episode-level MAE.}
  \label{fig:d1-memory-trajectory}
\end{figure}

\textbf{Progress ordering.}
Value-Order Correlation (VOC), introduced by GVL~\cite{ma2025gvl}, measures rank agreement between predicted progress and chronological frame order.  This reference order assumes that progress increases with time, so VOC can penalize correctly predicted decreases during execution mistakes. We therefore use Spearman's rank correlation coefficient to compare predicted progress with our stage- and error-aware targets.
Let $\hat r_{e,t}$ and $\tilde r_{e,t}$ be the ranks of $\hat y_{e,t}$ and
$\tilde y_{e,t}$ within episode $e$, assigning tied values their average rank.
Both rank sequences have average $m_e=(|\mathcal F_e|+1)/2$. We compute
\begin{align}
  \rho_{\mathrm S,e}
  &= \frac{\sum_{t\in\mathcal F_e}
      (\hat r_{e,t}-m_e)(\tilde r_{e,t}-m_e)}
     {\sqrt{\sum_{t\in\mathcal F_e}(\hat r_{e,t}-m_e)^2}
      \sqrt{\sum_{t\in\mathcal F_e}(\tilde r_{e,t}-m_e)^2}},
      \nonumber\\
  \rho_{\mathrm S}
  &= \frac{1}{|\mathcal E|}\sum_{e\in\mathcal E}
     \rho_{\mathrm S,e}.
  \label{eq:rank-metric}
\end{align}
Higher $\rho_{\mathrm S}$ indicates closer agreement with the reference ordering. For strictly increasing targets, this reduces to rank-based VOC under the same tie convention.

\textbf{Temporal smoothness.}
We measure local changes in progress increments using the second difference
$\Delta^2 y_{e,t}=y_{e,t+1}-2y_{e,t}+y_{e,t-1}$.
Let $\mathcal T_e$ contain the center indices of consecutive three-frame
windows in $\mathcal F_e$ that do not cross a temporal reset.
Prediction curvature $S$ and curvature error $J$ are
\begin{align}
  S &= \frac{1}{|\mathcal E|}\sum_{e\in\mathcal E}
       \frac{1}{|\mathcal T_e|}\sum_{t\in\mathcal T_e}
       \left|\Delta^2\hat y_{e,t}\right|, \nonumber\\
  J &= \frac{1}{|\mathcal E|}\sum_{e\in\mathcal E}
       \frac{1}{|\mathcal T_e|}\sum_{t\in\mathcal T_e}
       \left|\Delta^2\hat y_{e,t}-\Delta^2\tilde y_{e,t}\right|.
  \label{eq:temporal-metrics}
\end{align}
Lower $S$ indicates smoother predictions, while lower $J$ indicates closer
agreement with reference changes in slope.

\subsection{Ablation Studies and Comparisons}

We compare five ablations (A1--A5) and the SARM baseline to
address \textbf{Q1--Q4}. The full \method{} model uses an eight-slot recurrent
visual memory, a frozen VLM backbone, and causal peak selection.
A1--A4 are evaluated on parcel handling (\textit{D1}).
A5 and SARM are evaluated on both \textit{D1} and medicine-bottle
scanning and storage (\textit{D2}).

\textbf{VLM fine-tuning (Q2).}
\textbf{A1} unfreezes the VLM backbone during value-model training.
Comparing A1 with the frozen-backbone \method{} configuration evaluates
whether adapting the pretrained vision--language representations improves
progress estimation.

\textbf{Memory design (Q1).}
A2--A4 replace recurrent memory with fixed observation windows
(Fig.~\ref{fig:temporal-context}). Each variant fuses four multi-view
observation sets, sampled at offsets $(-3,-2,-1,0)$ for \textbf{A2},
$(-30,-10,-3,0)$ for \textbf{A3}, and $(-90,-30,-10,0)$ for \textbf{A4}.
Offsets are measured in frames, with $0$ denoting the current frame.
These variants keep the number of sampled observations fixed while extending
the temporal span, allowing us to compare recurrent memory with explicit
histories of different lengths.

\textbf{Peak selection (Q3).}
\textbf{A5} disables causal peak selection and instead uses the median of
the predicted categorical distribution at each frame. Both readouts use the
same checkpoint and categorical predictions, with recurrent memory retained.
This comparison isolates the contribution of peak selection to progress
accuracy and temporal consistency.

\textbf{Comparison with SARM (Q4).}
We compare \method{} with \textbf{SARM}~\cite{chen2026sarm}, a
state-of-the-art progress estimation method. SARM is trained using the same stage- and error-aware
progress targets as \method{}. We report $A$, $\rho_{\mathrm S}$, $S$, and $J$
for both the ablations and the baseline comparison.

\input{sections/results_discussion}

%% file: sections/results_discussion.tex
\subsection{Results and Discussion}
\label{sec:results}


Table~\ref{tab:ablation-comparison} reports episode-averaged metrics: $A$ is shown in units of $10^{-2}$,
$S$ and $J$ in units of $10^{-3}$, and Spearman's $\rho_{\mathrm S}$ on its
native $[-1,1]$ scale.

\begin{table}[t]
  \centering
  \caption{Ablations and comparison with SARM on D1 and D2.
  Best values before rounding are bold.}
  \label{tab:ablation-comparison}
  \footnotesize
  \setlength{\tabcolsep}{4pt}
  \begin{tabular}{@{}llrrrr@{}}
    \toprule
    Dataset & Method & $A\downarrow$ & $\rho_{\mathrm S}\uparrow$ & $S\downarrow$ & $J\downarrow$ \\
    \midrule
    \multirow{7}{*}{D1} & \method{} & 1.100 & \textbf{0.999} & 4.181 & 4.188 \\
                       & A1 & 1.150 & 0.998 & 4.821 & 4.826 \\
                       & A2 & 1.490 & 0.986 & 6.478 & 6.483 \\
                       & A3 & 1.203 & 0.996 & 5.456 & 5.460 \\
                       & A4 & 1.137 & 0.997 & 4.871 & 4.876 \\
                       & A5 & \textbf{1.087} & 0.999 & \textbf{3.915} & \textbf{3.920} \\
                       & SARM & 1.767 & 0.988 & 12.882 & 12.883 \\
    \midrule
    \multirow{3}{*}{D2} & \method{} & \textbf{3.307} & 0.959 & \textbf{5.660} & \textbf{5.665} \\
                       & A5 & 6.044 & 0.870 & 6.992 & 6.994 \\
                       & SARM & 3.470 & \textbf{0.978} & 16.585 & 16.586 \\
    \bottomrule
  \end{tabular}
\end{table}

\paragraph{Q1: Recurrent Memory versus Explicit Windows}

Extending the observation window from A2 to A4 improves all four metrics
on \textit{D1} (Table~\ref{tab:ablation-comparison}). Since these variants retain
the same number of observations, the trend suggests that earlier task states
provide useful context for interpreting the current scene. Recurrent memory
further improves on A4, reducing $A$ by 3.3\% and $J$ by 14.1\%.
The larger improvement in curvature error suggests that maintaining a
continuously updated history is particularly useful for temporal consistency.
Fig.~\ref{fig:d1-memory-trajectory} illustrates this benefit: \method{} tracks
the nonzero initial progress, while A2--A4 initially underestimate it.

\paragraph{Q2: Frozen versus Fine-Tuned VLM}

Freezing the VLM improves all four metrics on \textit{D1}
(Table~\ref{tab:ablation-comparison}) and reduces training time from 66.2 to
40.3 hours on eight H200 GPUs. This suggests that the pretrained
representations already contain useful cues for progress estimation, which
can be exploited by adapting the temporal memory and value readout.
The additional flexibility of backbone fine-tuning does not translate into
better test performance under the current training setup, making a frozen
VLM both an effective and computationally efficient choice.


\paragraph{Q3: Causal Peak Selection}

The effect of peak selection differs across the two tasks
(Table~\ref{tab:ablation-comparison}). On \textit{D1}, where only one of 196 test
episodes contains an annotated error, peak selection offers little benefit;
the median readout yields slightly lower $A$, $S$, and $J$. In contrast, on
the 120-episode \textit{D2} test set, which contains 25 error
episodes, peak selection reduces $A$ by 45.3\% and raises
$\rho_{\mathrm S}$ from 0.870 to 0.959, while also lowering $S$ and $J$.
This contrast is consistent with peak selection being particularly useful
when execution errors make progress ambiguous. A per-frame median can jump
as probability mass shifts between competing progress hypotheses; associating
supported peaks across time favors continuity while still allowing progress
to decrease. The joint improvement in accuracy, ordering, and curvature error
on \textit{D2} supports the value of temporal continuity for tracking progress
through execution setbacks.

\paragraph{Q4: Comparison with SARM}

\method{} achieves lower $A$, $S$, and $J$ than SARM on both tasks, with
curvature error reduced by over 65\% in each case
(Table~\ref{tab:ablation-comparison}). The reductions in both absolute and
curvature error indicate that smoother predictions also capture progress
magnitude and changes in its rate more accurately. Progress ordering also
improves on \textit{D1}, while SARM achieves higher $\rho_{\mathrm S}$ on
\textit{D2}. Fig.~\ref{fig:d2-baseline-trajectory} provides a representative
example: \method{} follows the target decrease during an error and the
subsequent recovery more closely than SARM. These gains under shared progress
supervision support the combination of temporal memory, distributional
prediction, and causal peak selection for accurate and temporally consistent
progress monitoring.

%% file: sections/conclusion.tex
\section{Conclusion and Future Work}

We introduced ReSTEP, a real-robot dataset that captures task progression,
execution errors, and recovery through detailed stage and error annotations.
By making these behaviors available for training and evaluation, ReSTEP
provides an important benchmark for future work on progress monitoring with
error awareness.

We also proposed \method{}, which combines temporal memory, distributional
prediction, and causal peak selection to track task progress through visual
ambiguity and execution setbacks. Experiments on both ReSTEP tasks demonstrate
more accurate and temporally consistent progress estimates than the
state-of-the-art SARM baseline.

Current methods rely on task-specific annotations and have been evaluated on only a single robotic platform across two task types. Future work will focus on expanding the scope of tasks and robotic platforms, reducing annotation requirements, and leveraging progress feedback to optimize control strategies and enable robot self-recovery for self-evolution. The current visual memory approach serves as a foundational prototype for video-based demonstration and in-context learning.